\documentclass[final,5p,times,twocolumn, authoryear]{elsarticle}

\usepackage{algorithm} 
\usepackage{algpseudocode} 
\usepackage{array}
\usepackage{amssymb}
\usepackage{amsmath}
\usepackage{lineno}

\usepackage{color}

\newcolumntype{L}[1]{>{\raggedright\arraybackslash}p{#1}}
\usepackage{enumitem}

\journal{Ocean Engineering}

\begin{document}

\begin{frontmatter}

\title{Wake-Informed 3D Path Planning for Autonomous Underwater Vehicles Using A* and Neural Network Approximations}

\author[label1,label2]{Zachary Cooper-Baldock}
\author[label2]{Stephen R. Turnock}
\author[label1]{Karl Sammut}

\affiliation[label1]{organization={Centre for Defence Engineering, Research and Training (CDERT), Flinders University},
             city={Adelaide},
             country={Australia}}

\affiliation[label2]{organization={Maritime Engineering, Faculty of Engineering and Physical Sciences, University of Southampton},
             city={Southampton},
             country={United Kingdom}}

\begin{abstract} 

Autonomous Underwater Vehicles (AUVs) encounter significant energy, control and navigation challenges in complex underwater environments, particularly during close-proximity operations, such as launch and recovery (LAR), where fluid interactions and wake effects present additional navigational and energy challenges. Traditional path planning methods fail to incorporate these detailed wake structures, resulting in increased energy consumption, reduced control stability, and heightened safety risks. This paper presents a novel wake-informed, 3D path planning approach that fully integrates localized wake effects and global currents into the planning algorithm. Two variants of the A* algorithm -- a current-informed planner and a wake-informed planner -- are created to assess its validity and two separate neural network models are then trained, each designed to approximate one of the A* planner variants (current-informed and wake-informed respectively), enabling potential real time-application. Both the A* planners and NN models are evaluated using important metrics such as energy expenditure, path length, and encounters with high-velocity and turbulent regions. The results demonstrate a wake-informed A* planner consistently achieves the lowest energy expenditure and minimizes encounters with high-velocity regions, reducing energy consumption by up to 11.3\%. The neural network models are observed to offer computational speedup of 6 orders of magnitude, but exhibit 4.51--19.79\% higher energy expenditures and 9.81--24.38\% less optimal paths. These findings underscore the importance of incorporating detailed wake structures into traditional path planning algorithms and the benefits of neural network approximations to enhance energy efficiency and operational safety for AUVs in complex 3D domains.
\end{abstract}





\end{frontmatter}

\section{Introduction} 
\label{Introduction } 
Autonomous underwater vehicles (AUVs) operate in complex, rapidly changing environments, with limited sensory data available. This lack of sensory perception in a complex environment complicates the effective control of these vehicles from a guidance and navigation perspective. This is further compounded by the addition of multiple vehicles or the undertaking of challenging manoeuvres, such as underwater launch and recovery (LAR) of smaller platforms in the unsteady flow wake of a larger platform \citep{SUBSTEC, IEEE}. Whilst underwater and in operation, AUV platforms are free to move with 6 degrees of freedom and in doing so need to maintain effective surge, heave and sway control, in addition to pitch, yaw and roll \citep{X1}. 

The underwater environment is further complicated by communications challenges, current and dynamic flow effects and the size of the working environment \citep{X1}. To ensure safe and efficient navigational paths in the underwater domain, a variety of different path planning methods have been developed or adapted \citep{X2}. The majority of these methods seek to optimize travel time \citep{X3}, energy consumption \citep{X4}, and obstacle avoidance \citep{X5} and can be broadly classified into node-based, sampling-based, or bio-inspired, depending on their operational principles and original inspiration. 

Limited research has addressed path-planning where vehicle wake structures are included, specifically in 3D applications, opting to instead focus on global current effects \citep{X6}, geometric obstacle avoidance or simplified \citep{X7} flow field structures. The inclusion of flow structures is primarily limited to 2D \citep{X6, X7}. Works that have sought to investigate path planning within 3D flows for autonomous applications have stopped short of incorporating the 3D field directly within the model \citep{X8}, favoring the recommendation of navigational best practices instead. Other research focusing on unmanned aerial vehicle (UAV) navigation in turbulent 3D domains, using the A* algorithm, opted to completely avoid turbulent regions of the path to ensure flight safety \citep{X9}.

However, for close-proximity underwater operations between vehicles, such as LAR, the turbulent structures are large, strong, and hard to positionally approximate \citep{SUBSTEC, IEEE}. Due to this, when conducting close-proximity operations between vehicles, complete avoidance as used in other works \citep{X9} is not possible. Subsequently, this necessitates an intelligent, rapid, and computationally efficient means of navigating near or through wake structures borne from other vessels.

\subsection{Limitations in Current Research}

For underwater applications, and specifically those with two or more interacting vehicles, an understanding of the best path through a fully realized 3D field is important. These operations are common and include, but are not limited to, launch and recovery, formation flying and multi-vehicle underwater cable and pipe inspection operations. In these applications, the 3D flow field, including wake structures and shed hydrodynamics each of the vehicles may pose control, safety and energy efficiency effects that are important to account for \citep{Bhattacharyya2011}.

To date, limited research has sought to model these effects within a path planner, at a level sufficient to encapsulate the complex 3D hydrodynamic physics, including the wake structures of nearby vessels. Existing methods either simplify the flow field in relation to its dimensionality (3D to 2D \citep{Garau2005, Petres2007}), include only the current, avoiding turbulent regions altogether \citep{Pensado2024}, or use approximated hydrodynamic models \citep{GZhang2022, Yang2019, Zadeh2018, KarlSuggested, Yi2023}. Research focusing on UAV navigation in turbulent 3D aerial domains using the A* algorithm opted to completely avoid turbulent regions entirely to ensure flight safety \citep{Pensado2024}. However, when conducting close proximity operations between vehicles, such as docking, formation flying or collaborative tasks, avoiding these regions is not always a practical or possible option. Vehicles may be required to navigate in close proximity to achieve their goals. These existing models may be inadequate to capture the fluid dynamic impacts present in multi-vehicle operations, where strong 3D wake interactions can significantly impact AUV control, safety, and energy consumption \citep{SUBSTEC, IEEE, Bhattacharyya2011}. 

To investigate whether a detailed knowledge of these wake hydrodynamics is required, and the benefits of doing so, two A* path planners will be developed for comparison -- one in which only the global current is considered, and one where the detailed 3D wake structure is accounted for. Additionally, the use of machine learning based planners, trained on the A* trajectories, will also be investigated. Such an approach may enable the rapid prediction of feasible, energy-efficient paths without the need for computationally expensive A* searches, potentially bringing real-time 3D wake-informed navigation within reach.

\subsection{Contributions}
To address these scientific gaps, this paper makes the following primary contributions:

\begin{enumerate}
    \item Proposes a wake-informed A* path planning algorithm that integrates full 3D hydrodynamic data, including complex wake structures, to achieve more energy-efficient and safer AUV navigation. Bridging the gap between simplified flow models and the complex realities of 3D underwater environments is crucial for operations involving significant wake effects \citep{SUBSTEC, IEEE}.
    \item Introduces a machine learning framework trained on wake- and current-informed A* trajectories, enabling rapid real time path generation that approximates the near-optimal paths in real-time, levering existing techniques for ML informed planning \citep{Tai2017, Pfeiffer2017}.
    \item Develops a robust set of metrics to holistically evaluate energy consumption, path length, turbulence exposure, and computational efficiency, advancing the understanding of AUV path planning performance in complex 3D domains.

\end{enumerate}
To contextualize this work, Section \ref{Background} reviews the current state of path planning in the underwater domain and identifies key gaps that motivate the study. Section \ref{Computational Domain} introduces the computational domain and the specific AUV configurations considered. In Section \ref{Methodology}, the development of the wake- and energy-informed planning algorithm is presented, while Section \ref{Neural Network Design} discusses the neural network approach used to approximate the planners for real-time applications. The performance of both the A* planners and their neural network counterparts is evaluated in Section \ref{Results}, followed by a detailed discussion of the findings in Section \ref{Discussion}. Finally, Section \ref{Conclusions} concludes the paper and outlines potential directions for future research.

\section{Background} 
\label{Background} 

Effective path planning is crucial for Autonomous Underwater Vehicle (AUV) operations in complex environments, aiming to optimize metrics like travel time, energy consumption, and safety \citep{X3, X4, X5}. Table \ref{tab:LiteratureSummary} summarizes relevant path planning literature, highlighting the methods, considered environmental effects, dimensionality, and key limitations, particularly concerning the integration of detailed 3D hydrodynamics. Common approaches adapt terrestrial or aerial methods, broadly categorized as node-based (e.g. A* \citep{Garau2005}), sampling-based (e.g. RRT \citep{LaValle2006, RandomTrees}), or bio-inspired methods (e.g. Ant Colony Optimization \citep{AntColony, Alvarez2004}). The domain that AUVs operate in is complex, as they are affected by global current effects\citep{Kruger2007}, wake structures \citep{IEEE}, local obstacles that must be avoided and the hydrodynamics of other vessels \citep{IEEE}.

While numerous studies have integrated ocean currents to find time or energy-optimal routes \citep{Kruger2007, X2, Li2016}, these often rely on simplified 2D representations \citep{Garau2005, Petres2007} or extend currents to 3D using approximations that neglect the full dimensionality of the flow, omitting complex, localized hydrodynamic features like vehicle wakes \citep{Zadeh2018, KarlSuggested, Yi2023}. Research involving turbulent aerial domains has been observed to avoid the turbulent regions entirely \citep{Pensado2024}, an approach that is impractical or impossible for close-proximity underwater maneuvers such as launch and recovery \citep{SUBSTEC, IEEE}. More recently, machine learning techniques and specifically neural networks (NNs), have been applied to path planning tasks to capitalizing on computational speed and hardware requirement reductions \citep{Tai2017, Pfeiffer2017}. However, these are frequently applied to simplified 2D grids, and have a focus on obstacle avoidance rather than navigation of complex 3D flow fields.

\begin{table*}[!ht]
    \centering
    \small
    \caption{Summary of Relevant Path Planning Literature for Autonomous Maritime Vehicles.}
    \label{tab:LiteratureSummary}
    \begin{tabular}{L{2.5cm} L{1.5cm} L{3.5cm} L{1.5cm} L{3cm} L{3.5cm}}
        \hline
        Reference(s) & Method              & Included Effects                   & Dim. & Pros                                     & Limitations                         \\ \hline

        \citep{Garau2005}        & A* (Node-based)   & Spatially variable currents  & 2D   & Considers current effects on path cost.  & Limited to 2D; No wake structures.                      \vspace{0.3cm} \\ 
        \citep{Petres2007}       & Review (Multiple)        & Ocean currents (General focus)     & 2D   & Reviews various methods for current planning. & Highlights reliance on simplified 2D models.      \vspace{0.3cm}   \\ 
        \citep{X2}      & Evolutionary     & Ocean currents                   & 2D  & Optimize time/energy with currents.  & Limited to 2D; No wake structures.                        
    \vspace{0.3cm} \\
        \citep{Kruger2007}  & A* (Node-based)              & Estuary currents                   & 3D & Considers relative motion and energy   & Limited 3D flow model; No wakes           \vspace{0.3cm}              \\
        \citep{Li2016} & Optimization & Ocean currents                   & 3D & Optimize time/energy with currents.        & Uses simplified current fields; No detailed wakes. \vspace{0.3cm} \\ 
        \citep{Zadeh2018}        & Differential Evolution   & Ocean currents, 3D obstacles     & 3D   & Handles 3D obstacles and currents.       & Did not fully integrate complex hydrodynamic effects. \vspace{0.3cm} \\ 
        \citep{KarlSuggested}    & A*      & Extended 2D current data         & 3D   & Attempted 3D current representation.     & Simplified linear extension of 2D data; No wake.   \vspace{0.3cm}  \\
        \citep{Yi2023}           & Optimization             & Currents, Obstacles, Travel Dist.& 3D   & Detailed 3D constraints considered.      & Assumed simplified current flow; No wake structures.  \vspace{0.3cm}  \\ 
        \citep{Pensado2024}      & A*                       & Turbulence (Aerial UAV focus)      & 3D   & Addresses navigation in turbulence.     & Avoids turbulent regions entirely; Not wake-specific. \vspace{0.3cm} \\ 
        \citep{Tai2017, Pfeiffer2017} & Neural Networks (NN)   & Obstacles, Environment Features    & 2D/3D & Fast computation; Learning-based.        & Often applied to simpler env./grids; Not focused on detailed 3D wakes. \vspace{0.3cm} \\ 
        \textbf{This Work}      & \textbf{A* \& NN}& \textbf{Currents \& Detailed Wake} & \textbf{3D} & \textbf{Integrates full 3D wake data; Compares A*/NN.} & \textbf{Focuses on specific LAR scenario} \vspace{0.3cm} \\
        \hline
    \end{tabular}
\end{table*}

The literature summarized in Table \ref{tab:LiteratureSummary} indicates that while progress has been made in incorporating environmental factors like currents and navigating 3D spaces, a critical gap remains in path planning methodologies that fully integrate detailed, localized 3D hydrodynamic effects, specifically the complex wake structures generated by nearby vehicles. These wakes significantly impact AUV control, safety, and energy efficiency during close-proximity operations \citep{Bhattacharyya2011, SUBSTEC, IEEE}, yet are typically simplified or ignored. This research directly addresses this limitation by developing and evaluating path planners—both high-fidelity A* algorithms and computationally efficient neural network approximations—that explicitly incorporate detailed 3D wake data alongside global currents to improve AUV navigation performance in such challenging scenarios.

\section{Computational Domain} 
\label{Computational Domain}

Underwater LAR involves one vehicle advancing towards the other whilst both are fully immersed in the fluid domain. The immersion results in a three dimensional flow field, where the flow may move in the x, y or z directions. This leads to the presence of complex fluid structures that can impact the performance, stability and control of the involved vessels. In the domain as proposed in this investigation, it is assumed that a smaller underwater vehicle will be recovered within a larger underwater vehicle. The smaller vehicle will navigate upstream, from behind the stern of the steadily moving larger craft, during the approach with the goal of being recovered inside a payload bay located onboard the larger vessel. 

This method of recovery has already been proposed and hydrodynamically assessed \citep{SUBSTEC, IEEE}, but remains challenging due to the fluid structures present during the manoeuvre which are excluded in current path planners. The propeller race and vehicle wake of the lead vessel presents a complex navigational challenge to the approaching vehicle. Approaching with limited knowledge of the strong flow structure, instead accounting only for the current, can complicate attempts to make a safe recovery. Thus, a smart means of avoidance is required. This presents an optimal task by which to assess, benchmark and evaluate efficient methods for wake-informed path planning in 3D domains.

The LAR domain is modeled as a pseudo-static environment. The larger vehicle, an extra-large uncrewed underwater vehicle (XLUUV) is located within the center of the computational grid. This grid represents a 155 $\times$ 155 $\times$ by 155 meter physical domain. The larger vehicle to which the recovery is made is modeled to be 22 meters in length, 2.2 meters wide and 2.7 meters tall. This vehicle is equipped with an INSEAN E1619 propeller, based on the work of \citep{IEEE}. It is modeled with a central payload bay that is 5.5 meters long, 1.5 meters wide and 2.2 meters tall. It is assumed that for each planned trajectory, the XLUUV velocity and heading are constant. These manoeuvres take place at a modeled depth of 100m, with a constant current and no wave or free surface effects. 

\begin{figure}[!ht]
    \centering
    \includegraphics[width=0.95\linewidth]{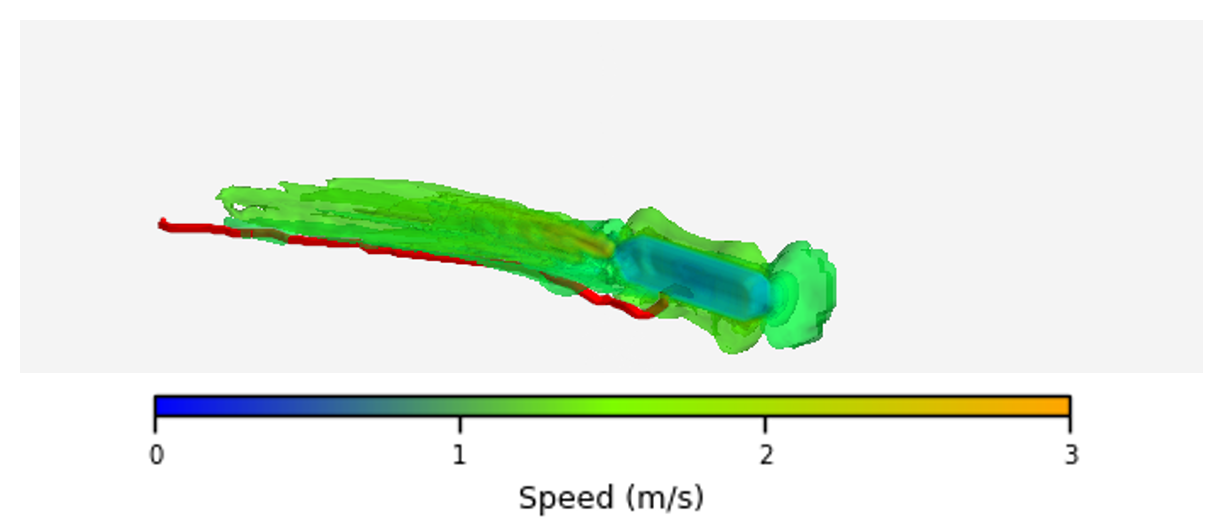}
    \caption{Side view of the optimal A* wake-informed path planner trajectory for a given flow state. The AUV follows the trajectory indicated in red, minimizing its interaction with the wake structure of the XLUUV, where the high speed flow is indicated in green. The XLUUV body geometry is visible in blue, through the transparent flow field in green. In this case, the AUV must navigate to the goal location within the XLUUVs' central payload bay, necessitating a transition through the boundary layer.}
    \label{fig:DomainSide}
\end{figure}

\begin{figure}[!ht]
    \centering
    \includegraphics[width=0.95\linewidth]{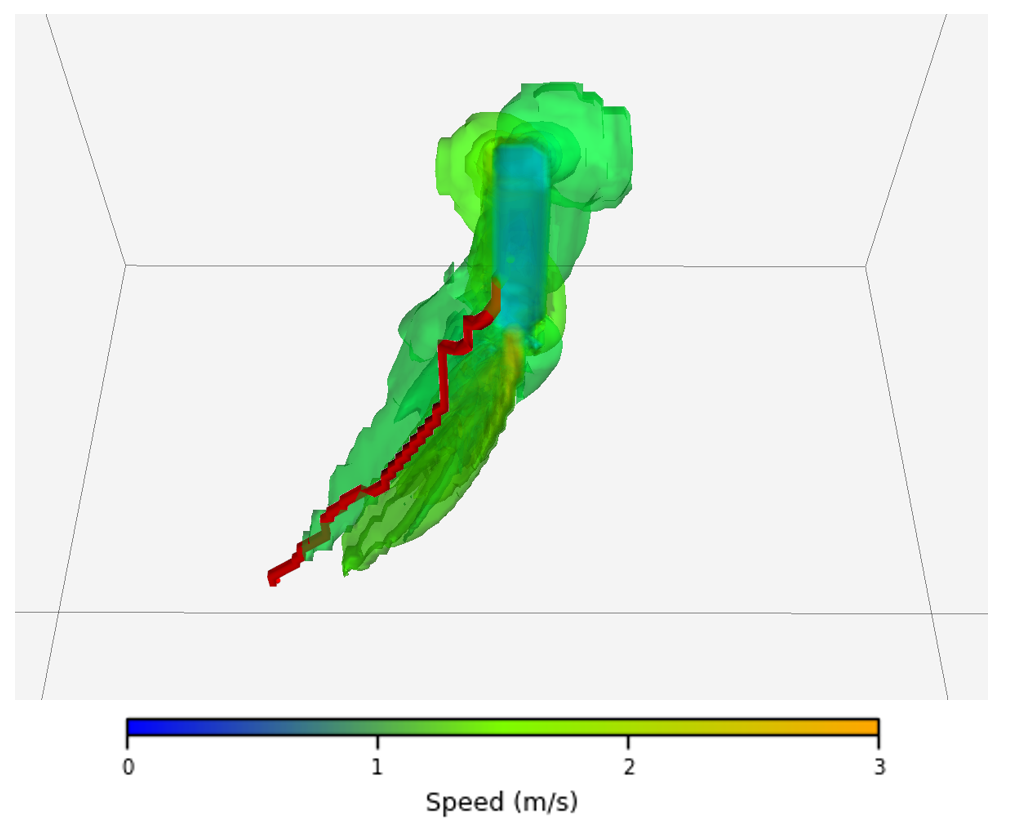}
    \caption{Bottom up view of the optimal A* wake-informed path planner trajectory for a given flow state. The AUV follows the trajectory indicated in red, minimizing its interaction with the wake structure of the XLUUV, where the high speed flow is indicated in green. The XLUUV body geometry is visible in blue, through the transparent flow field in green. In this case, the AUV must navigate to the goal location within the XLUUVs' central payload bay, necessitating a transition through the boundary layer.}
    \label{fig:DomainUnder}
\end{figure}

This payload bay is accessible from the bottom, mimicking the payload bay LAR as assessed by \citep{IEEE} and \citep{SUBSTEC}, which indicated such a maneuver is possible. This domain is contained within a dataset of 500 distinct flow conditions, as discussed in Section \ref{Path Generation}. The velocity magnitude contained within the ANSYS Fluent CFD data has been interpolated from the mesh nodes of each simulation onto the regularly spaced 3D grid structure used in the planner. This was undertaken using a nearest-neighbor approach between vertexes of the CFD data and the 3D grid structure as discussed in Section \ref{Path Planner Operation}. Example trajectories through the domain are provided in Fig. \ref{fig:DomainSide} and \ref{fig:DomainUnder} which show the larger vehicle, the wake structure and an example approach trajectory.

\section{Methodology} 
\label{Methodology}

\subsection{Path Planner Operation}
\label{Path Planner Operation}

The environment is first discretized into a three-dimensional grid $G \subset \mathbb{Z}^{3}$, where each cell (voxel) represents a discrete location in 3D space. The grid dimensions are defined to be ($[1,128],\ [1,128],\ [1,128]$) representing the 3D $128^{3}$ grid structure. Each individual cell in the grid corresponds to a node $n$. These nodes are connected to their neighbors based on a 26-connected grid, allowing movement to any adjacent voxel sharing a face, edge, or corner. For a node $n$, its set of neighbors $N(n)$ includes all adjacent nodes that are within the grid bounds and are traversable (not occupied by obstacles). The Euclidean distance between two nodes $n=(x_n, y_n, z_n)$ and $n'=(x_{n'}, y_{n'}, z_{n'})$ is denoted by $d(n, n')$ and is given by Eq. (\ref{eqn:EuclideanDistance}):

\begin{equation}
    \label{eqn:EuclideanDistance}
    d(n, n') = \sqrt{(x_{n'}-x_{n})^{2}+(y_{n'}-y_{n})^{2}+(z_{n'}-z_{n})^{2}}
\end{equation}

The cost of moving from node $n$ to an adjacent node $n'$ is defined based on the estimated energy required to overcome hydrodynamic drag force $F_{D}$ during that movement. This can be determined, as the velocity throughout the local environment is known (or estimated) within each cell of the grid $G$. The drag force ($F_{D}$) acting on the AUV is determined via Eq. (\ref{eqn:DragForce}). The AUV particulars and values used to enable accurate drag calculation for Eq. (\ref{eqn:DragForce}) are provided in \ref{app2}. 

\begin{equation}
    \label{eqn:DragForce}
    F_{D}=\frac{1}{2} \rho v^{2} C_{D} A
\end{equation}
In Eq. (\ref{eqn:DragForce}) $\rho$ denotes the fluid density, $v$ denotes the magnitude of the relative velocity between the vehicle and the fluid, $C_D$ denotes the coefficient of drag for the AUV, and $A$ denotes its cross-sectional area. The drag can be used in conjunction with the Euclidean distance to determine the energy cost associated with moving to $n'$. To calculate this cost of traversal ($n$ to $n'$), we evaluate the drag force using the fluid velocity conditions at the destination node $n'$, denoted $F_D(n')$. The subsequent energy $E_D(n, n')$ required to overcome this drag, over the distance $d(n, n')$, is approximated as by Eq. (\ref{eqn:DragEnergy}).

\begin{equation}
    \label{eqn:DragEnergy}
    E_{D}(n, n') = F_{D}(n') \times d(n, n')
\end{equation}
\noindent where $F_D(n') = \frac{1}{2} \rho v(n')^2 C_D A$, using the relative fluid velocity $v(n')$ estimated at node $n'$.

The energy cost $E_{D}(n, n')$ is then used to define the movement cost $c(n, n')$, also known as the edge weight, for traversing from node $n$ to node $n'$. A weighting factor $\omega_{1}$ (typically set to 1 unless scaling is desired) can be included as denoted in Eq. (\ref{eqn:NodeCost}) which states the movement cost as a function of the energy ($E_{D}$) which is comprised of both the fluid forces ($F_{D}$) and the movement distance ($d$).

\begin{equation}
    \label{eqn:NodeCost}
    c(n, n') = \omega_{1} \cdot E_{D}(n, n') = \omega_{1} \cdot F_{D}(n') \cdot d(n, n')
\end{equation}

The A* algorithm searches for a path by maintaining the cost of the path found so far from the start node $n_0$ to the current node $n_{curr}$. This is known as the $g$-score, denoted by $g(n_{curr})$, and is calculated by summing the edge costs along the path given by $P = [n_0, n_1, \dots, n_{curr}]$. The equation for this is provided in Eq. (\ref{eqn:CostFunction}).

\begin{equation}
    \label{eqn:CostFunction}
    g(n_{curr}) = \sum_{i=1}^{curr} c(n_{i-1}, n_{i})
\end{equation}

The path planning problem is formulated as an optimization task. The design variable is the path $P$, represented as a sequence of connected nodes $[n_0, n_1, \dots, n_k]$ from a given start node $n_0$ to a specified goal node $n_k$. The objective is to find the optimal path $P^*$ that minimizes the total cumulative cost, $g(n_k)$, required to reach the goal $n_k$, as stated by Eq. (\ref{eqn:PathFunction}).

\begin{equation}
    \label{eqn:PathFunction}
     P^* = \underset{P \in \mathcal{P}(n_0, n_k)}{\operatorname{arg\,min}} \ g(n_k) = \underset{P = [n_0, ..., n_k]}{\operatorname{arg\,min}} \left( \sum_{i=1}^{k} c(n_{i-1}, n_{i}) \right)
\end{equation}

\noindent where $\mathcal{P}(n_0, n_k)$ is the set of all possible valid paths from $n_0$ to $n_k$.These valid paths are assessed via a heuristic function. The A* algorithm utilizes the heuristic function, $h(n_{curr})$, to estimate the minimum remaining cost from the current node $n_{curr}$ to the goal node $n_k$. This estimate guides the search efficiently towards the goal. In this work, the heuristic is defined as the product of the straight-line Euclidean distance $d(n_{curr}, n_k)$ between the current node ($n_{curr}$) and the goal ($n_k$). This is provided in Eq. (\ref{eqn:SearchHeuristic}).

\begin{equation}
    \label{eqn:SearchHeuristic}
    h(n_{curr}) = \overline{E_{D}} \times d(n_{curr}, n_k)
\end{equation}

Here, $\overline{E_{D}}$ represents the global average of the energy term, which consists of the weighted drag force term, $\omega_1 \cdot F_D(n')$ (derived from Eq. \ref{eqn:NodeCost}). This heuristic combines the geometric distance to the goal with an overall estimate of the environmental energy resistance. Crucially, this heuristic is admissible because the average energy cost $\overline{E_D}$ will not overestimate the actual minimum energy cost per unit distance along any specific optimal path segment. Since the actual cost on the optimal path from $n$ to $n_k$ involves traversing specific cells with potentially lower-than-average or higher-than-average costs, using the global average to guide the search prevents systematic overestimation of the remaining cost. Admissibility is a required property for A* to guarantee finding the minimum cost path \citep{Hart1968}. While simple, this heuristic provides a computationally inexpensive way to focus the search, balancing distance and the energy landscape of the grid $G$.

The A* algorithm explores nodes based on the priority function $f(n) = g(n) + h(n)$. It maintains an 'open set' (priority queue) of nodes to be evaluated, ordered by their $f$-score, and a 'closed set' of nodes already evaluated. Starting with $n_0$, it iteratively selects the node $n$ with the lowest $f$-score from the open set, expands its neighbors $n'$, calculates their tentative $g$-scores and $f$-scores, and updates them if a cheaper path is found. The process continues until the goal node $n_k$ is selected for expansion, at which point the optimal path is reconstructed by backtracking from the goal.

To assess the importance of knowing the wake structures, two variants of the planner will be assessed. The first variant, $P^{*}_C$ will only be informed of the current within the computational domain. The second variant, $P^{*}_W$, will be informed of both the current and the vehicle's wake structure. $P^{*}_W$ will be applied to the flow field across a range of wake fields behind a large underwater vehicle. The current and wake dataset was generated from prior work by the authors and is comprised of high fidelity computational fluid dynamics (CFD) data interpolated onto the discrete grid $G \subset \mathbb{Z}^{3}$ \citep{IEEE} \citep{SUBSTEC}. More details about this are provided in \ref{app1}. The data contains a local current from 0.10 m/s to 5.00 m/s in increments of 0.10 m/s. Additionally, a strong propeller wake is included, with wake structures at 0° to 60° degrees of separation, simulating a turn of the vehicle in question. This was done via the simulation of an INSEAN E1619 propeller. The wake free velocity field ($v_C$) containing only the current, will be the same field, with the wake structure’s components filtered out and only global current retained.

\subsection{Graphical Depiction}

\begin{figure}[!ht]
    \centering
    \includegraphics[width=0.65\linewidth]{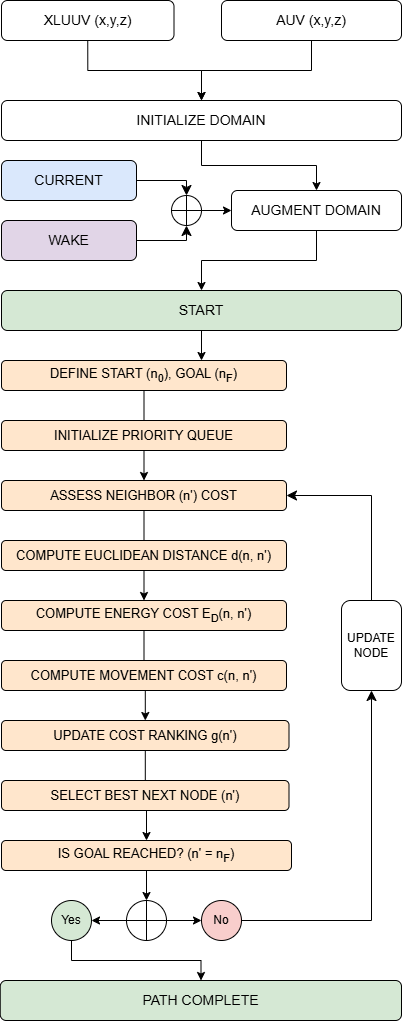}
    \caption{Operation of the A* planner architecture. Domain is initialized with the position of the XLUUV and AUV relative to each other. Method of flow data initialization (current only or wake) indicated in blue and purple respectively. Operation of the A* planner node selection indicated in orange.}
    \label{fig:PlannerOperation}
\end{figure}

Fig. \ref{fig:PlannerOperation} details the A* path planner methodology. The positions of the XLUUV and AUV (x, y, z) are used to discretize a 3D domain in the $128^{3}$ structure as discussed. This grid structure is then augmented with either the current or wake flow data as denoted (blue and purple respectively). Following this, the path planner is started, with the start and goal locations ($n_0, n_k$) defined. A priority queue is initialized before the assessment process for the neighboring nodes ($n'$) using the search heuristic ($h$) is undertaken. This process is iterated, with the best node selected based on the cost until the goal location is reached. At this point, the planner is stopped, as a trajectory has been generated.

\subsection{Path Generation}
\label{Path Generation}

To compare the effectiveness of the wake-informed A* planner and the current-informed A* planner, many trajectories needed to be generated for assessment. This was achieved through the use of the CFD dataset (\ref{app1}). This dataset contained 500 distinct flow fields, each of which were characterized by a different speed and angle pairing, and resulted in a different wake flow field behind the XLUUV. 

\begin{figure}[!ht]
    \centering
    \includegraphics[width=0.95\linewidth]{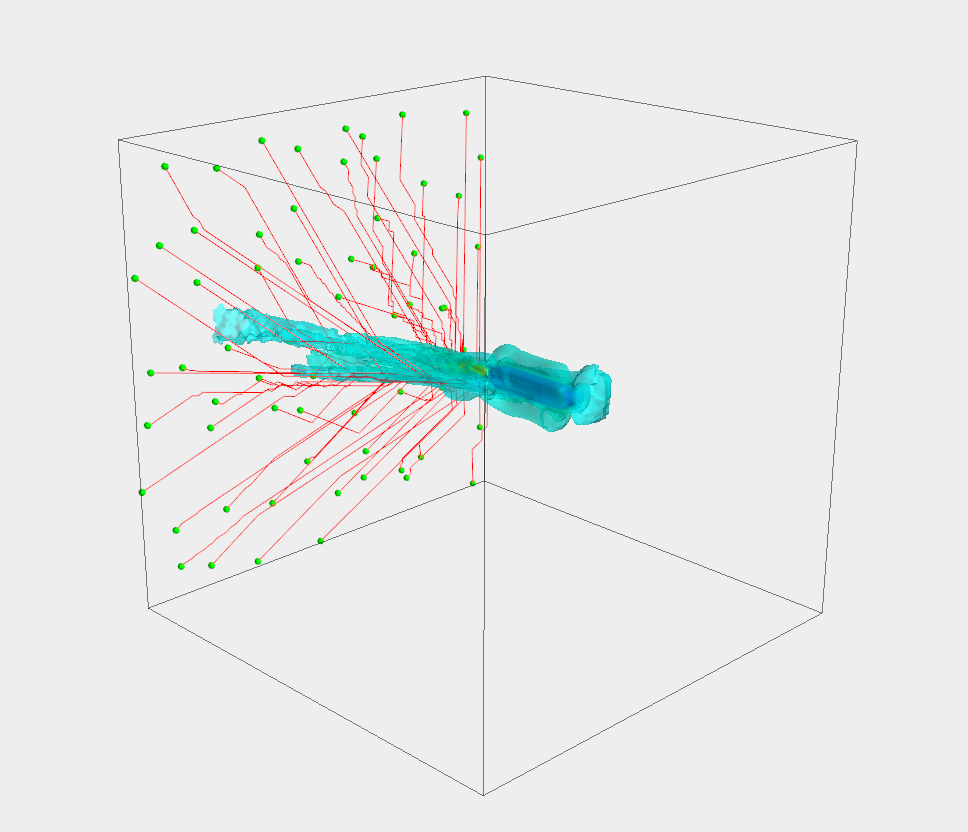}
    \caption{Example path generation of multiple A* trajectories shown. Each trajectory is indicated in red, with the goal of avoiding the wake flow structure (indicated in blue). Domain is shown from a front 3/4 side view. The starting location of each AUV trajectory is indicated by the green sphere. The trajectory goal location is the center of the payload bay within the XLUUV, including the final stage recovery, through the boundary layer. The XLUUV body geometry is contained within the blue wake flow.}
    \label{fig:DomainSidePaths}
\end{figure}

For each of these 500 instances, the location of the AUV was randomly generated on the rear boundary of the 3D domain (depicted in Fig. \ref{fig:DomainSidePaths}). If the current-informed planner was being assessed, the grid velocities $G(v)$ would be overwritten with the mean velocity value ($\overline{v}$) in all cells, resulting in a domain comprising of only the global current. If the wake-planner was being assessed, then the wake and detailed velocity structure of the CFD data would be retained. 

The randomized start points of the AUV, of which 36 were generated for each of the 500 fields, were held consistent for both the current- and wake-informed planners. From these starting positions, the path planning process (outlined in Fig. \ref{fig:PlannerOperation}) would be followed, until the AUV had been navigated to the center of the payload bay onboard the XLUUV (the goal $n_k$). Using 36 trajectories for each of the 500 flow conditions resulted in 18,000 paths being determined for the current- and wake-informed planners respectively. 

An example instance, of 1 of the 500 distinct flow fields, is provided in Fig. \ref{fig:DomainSidePaths}. In this image, the 36 paths generated for the wake-informed A* planner are shown. The 36 starting locations of the AUV on the rear boundary are denoted by the green spheres. The trajectories towards and into the XLUUV are denoted by the red lines, and the wake velocity data ($v_{wake}$) is provided in blue, with the XLUUV located in the center of the wake field.

\subsection{Path Assessment Metrics}
\label{Path Assessment Metrics}

The energy capacity onboard an AUV is extremely limited. This is due to their small physical size and inability to house large battery systems \citep{SUBSTEC}. Due to the limited energy budgets, there are several important performance metrics that should be assessed for any trajectory.

A successful trajectory should be energy efficient, short in physical length and also minimize the traversal of adverse high velocity or turbulent regions of the domain $G$. To assess these metrics, several performance metrics have been designed. These are the method to determine the energy cost of the path ($E$), the length of the path ($L$), the number of high velocity cells traversed $N_{\text{high-velocity}}(P)$) and the number of turbulent cells traversed ($N_{\text{high-velocity}}(P)$). 

To make these assessments, each generated path, $P$, is decomposed into a set of $N$ nodes such that:

\begin{equation}
    \label{eqn:PathDecomp}
   P = \{ n_1, n_2, n_3, \dots, n_N \} 
\end{equation}

Each individual node ($n_i$) in the set corresponds to a 3-dimensional coordinate $(x_i, y_i, z_i)$ in the grid. The velocity at node $n_i$ is $v(n_i)$. $\tilde{v}$ denotes the median velocity of the entire grid $G$, and $\sigma_v$ denotes the standard deviation of the velocities in the grid. Crucially, whilst the current-informed planner operated without knowledge of the wake flow in $G$, it will be assessed in a domain containing the wake flow, in order to assess the impact of the wake on the metrics identified above and the impact of accounting for the wake in the trajectory planning stage. $\sigma_v$ is used to identify high velocity cells, where we define a high velocity cell to be 1 standard deviation or more from the median velocity ($\tilde{v}$). Where an individual node ($n_i$) encounters a cell that is more than 1 standard deviation from the velocity field's mean value in the grid $G$, it is recorded, indicating the traversal of a high velocity cell. This threshold is given by Eq. \ref{eqn:v_threshold}

\begin{equation}
    \label{eqn:v_threshold}
    v_{\text{threshold}} = \tilde{v} + \sigma_v
\end{equation}

The number of cells along the trajectory that exceed this threshold is denoted by Eq. \ref{eqn:PathHighVelocity}. This equation tracks the number of high velocity cells via $N_{\text{high-velocity}}(P)$ and is calculated via Eq. (\ref{eqn:PathHighVelocity}), where the cutoff threshold for a high velocity cell is denoted by the indicator function in Eq. (\ref{eqn:PathVelocityCutOff}) and given via $\delta_{\text{V}}$ as mentioned.

\begin{equation}
\label{eqn:PathHighVelocity}
    N_{\text{high-velocity}}(P) = \sum_{i=1}^{N} \delta_{\text{V}}(n_i)  
\end{equation}

\begin{equation}
\label{eqn:PathVelocityCutOff}
    \delta_{\text{V}}(n_i) =
    \begin{cases}
    1, & \text{if } v(n_i) \geq v_{\text{threshold}} \\
    0, & \text{otherwise}
    \end{cases}
\end{equation}

To assess the degree of turbulence encountered over the trajectory, a similar method is applied. Where the velocity is found to change at a given node (via $v'(n_i)$), it is summed, and thus tracks the number of velocity fluctuations recorded over the path of the trajectory (through $N_{\text{turbulent}}(P)$). The CFD flow field data, determined via ANSYS Fluent, may have slight fluctuations due to numerical precision, which can result in small changes in velocity despite uniform freestream velocity conditions. To stop these cells from being incorrectly tracked, a small offset is applied $(\varepsilon = 0.005)$ ensuring that the velocity fluctuation, sufficient to be recorded as a turbulent cell transition, must be more than 0.5\% of $v$. The two equations to enable this tracking are Eq. (\ref{eqn:PathTurbulence}) and (\ref{eqn:PathTurbulenceCutOff}) respectively.

\begin{equation}
\label{eqn:PathTurbulence}
   N_{\text{turbulent}}(P) = \sum_{i=1}^{N} \delta_{\text{T}}(n_i) 
\end{equation}

\begin{equation}
\label{eqn:PathTurbulenceCutOff}
    \delta_{\text{T}}(n_i) =
    \begin{cases}
    1, & \text{if } v'(n_i) \geq \varepsilon \\
    0, & \text{otherwise}
    \end{cases}
\end{equation}

To determine the energy cost of a path, the drag force ($F_D$) as denoted by Eq. \ref{eqn:DragForce} and distance ($d$) denoted by Eq. \ref{eqn:EuclideanDistance} of each node ($n_i$) and the next ($n_{i+1}$) are summed along the trajectory via the application of \ref{eqn:PathEnergy}. This yields the energy cost of a path $E(P)$ in joules.

\begin{equation}
\label{eqn:PathEnergy}
    E(P) = \sum_{i=1}^{N-1} F_D(n_i, n_{i+1}) \times d(n_i, n_{i+1})    
\end{equation}

The total path length ($L$) along the set of nodes in  $P$, denoted $L(P)$, is the sum of the distances between consecutive nodes where $d_i$ is as previously defined by \ref{eqn:EuclideanDistance}. 

\begin{equation}
\label{eqn:PathLength}
   L(P) = \sum_{i=1}^{N-1} d(n_i, n_{i+1}) 
\end{equation}


\section{Neural Network Design} 
\label{Neural Network Design}

While the A* algorithm, when using an admissible heuristic as employed here (see Section 4.1), is guaranteed to find the path with the minimum cumulative cost on the discretized grid graph \citep{Hart1968}, this calculation requires substantial computing resources. Such resources are typically unavailable on an AUV with it's limited computational hardware and energy budget, rendering it either impractical or impossible to undertake such planning in real time.

Neural networks (NNs), in contrast, are capable of learning complex mappings from input features to outputs. This makes them ideal candidates to approximate the function of both the wake- and current-informed path planning A* algorithms \citep{LeCun2015}. Beneficially, NNs are significantly less computationally intensive (once trained) and offer the ability to rapidly generate and update the trajectories based on a new inputs, unlike the recalculations and computational resources required for traditional A*. 

To investigate the potential for real-time path generation, two distinct neural network (NN) models were developed and trained. Each NN was specifically designed to approximate the output of one of the A* planners: one network learns from the trajectories generated by the current-informed A* ($C.I._{A*}$) and the other learns from the wake-informed A* ($W.I._{A*}$) trajectories. This separation allows each network to specialize in mimicking the behavior of its corresponding A* algorithm, which operates on different environmental input data in the domain $G$ (current-only vs. current and wake).

\subsection{Dataset Preparation} 

To train the networks to predict the trajectories of the current and wake-informed A* planners, a dataset of their trajectories will be used. Each dataset consists of 18,000 trajectories generated by the wake-informed and current-informed A* planners counterparts, as described in Section~\ref{Path Generation}. Each trajectory comprises up to 130 waypoints, with each waypoint represented by its three-dimensional coordinates $(x_i, y_i, z_i)$. The input features for the neural network include the vehicle's starting position $(x_{\text{start}}, y_{\text{start}}, z_{\text{start}})$, goal position $(x_{\text{goal}}, y_{\text{goal}}, z_{\text{goal}})$, flow speed $v \in [0.1, 5.0]$ m/s, and flow angle $\theta \in [0^\circ, 60^\circ]$. The network output is a vector of sequential waypoint coordinates representing the predicted trajectory. Due to varying path lengths, all trajectories were padded to a fixed length of 130 waypoints, and a mask was applied during training to handle the variable lengths, that may be shorter than this length of 130 cells.

\subsection{Model Architecture} 

\begin{figure}[!ht]
    \centering
    \includegraphics[width=0.5\linewidth]{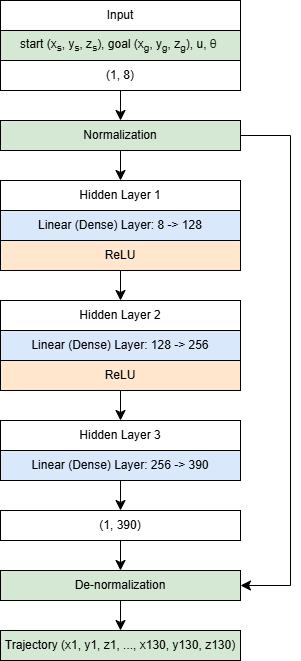}
    \caption{Neural network topology. Input to the network is the start location ($x_s, y_s, z_s$), goal location ($x_g, y_g, z_g$), flow speed ($u$) label and angle of attack label ($\theta$). The input vector has dimension (1, 8). This is then progressively increased by the linear layers to (1, 390), which is decomposed to a trajectory through ($[x_1,y_1,z_1],...,[x_{130}, y_{130}, z_{130}]$). This is then compared to the trajectories of the respective planner the NN is attempting to replicate ($W.I._{A*}$ or $C.I._{A*}$), before the loss is computed and backpropagated through the network.}
    \label{fig:NNarchitecture}
\end{figure}

Both neural network architectures are fully connected feedforward networks, designed to regress the sequence of waypoint coordinates directly from the input features. The architecture consists of an input layer, two hidden layers and an output layer as detailed in Fig. \ref{fig:NNarchitecture}.

Further details are provided in Table \ref{tab:NN_Architecture_Summary} in \ref{app3}. The choice of layer sizes was determined empirically to balance model capacity and computational efficiency. ReLU activation functions were used to introduce non-linearity and mitigate the vanishing gradient problem \citep{Glorot2011}. A fully connected architecture was chosen for its simplicity and to serve as a baseline, ensuring that future studies can easily compare advanced architectures (e.g., RNNs, Transformers) against this initial model.

\subsection{Training Procedure}

Both input features and output waypoint coordinates were normalized using the standard score as represented by Eq. (\ref{eqn:StandardScore}).

\begin{equation} 
    \label{eqn:StandardScore}
    z = \frac{x - \mu}{\sigma} 
\end{equation}

\noindent where $x$ is the feature value, $\mu$ is the mean, and $\sigma$ is the standard deviation computed from the training set. The dataset was then split into training and validation sets using an 80/20 split, resulting in 14,400 training samples and 3,600 validation samples. The split was stratified to ensure a uniform distribution of flow speeds and angles in both sets. A masked mean squared error (MSE) loss function was employed to score the ability of the two networks to predict wake-informed or current-informed trajectories. To determine the loss for a given path, the MSE loss was computed between the path generated by the neural network and the corresponding known path from the training set. This loss was then back-propagated through the network. The loss function is given by Eq. (\ref{eqn:MaskedMSE}).

\begin{equation} 
    \label{eqn:MaskedMSE}
    \mathcal{L} = \frac{\sum_{i=1}^{N} m_i (y_i - \hat{y}i)^2}{\sum{i=1}^{N} m_i} 
\end{equation}

\noindent where $y_i$ and $\hat{y}_i$ are the true and predicted waypoint coordinates, respectively, $m_i$ is the mask indicating valid waypoints ($m_i = 1$) or padded positions ($m_i = 0$), due to the variable trajectory length consisting of $N$ way-point coordinates in the path.

The network was trained using the Adam optimizer \citep{Kingma2015} with a learning rate of $1 \times 10^{-4}$. Adam was chosen for its adaptive learning rate capabilities and robustness in training deep neural networks. The batch size was set to 64, and the model was trained for 50 epochs. Weight initialization was performed using the Xavier initialization method to ensure stable gradients at the start of training \citep{Glorot2010}.

\subsection{Regularization Techniques} 

To prevent overfitting, three regularization techniques were implemented: \textbf{(1)} Early Stopping: The validation loss was monitored, and training was halted if the loss did not improve for a set number of epochs (10). \textbf{(2)} Dropout: Although not included in the final model, dropout layers were experimented with during hyperparameter tuning but did not yield significant improvements. \textbf{(3)} L2 Regularization: L2 weight decay was applied implicitly via the Adam optimizer's default settings. 

\subsection{Evaluation Metrics} 

The model's performance was evaluated using the root mean squared error (RMSE) between the predicted and true waypoint coordinates on the validation set only. Additionally, the trajectories generated by the neural network were assessed using the same metrics as the A* planners, including total energy consumption, path length, number of high-velocity cells, and number of disparity cells, as defined in Section~\ref{Path Assessment Metrics} and presented in Section~\ref{Results}.

\subsection{Implementation Details} 

The neural network was implemented using PyTorch \citep{Paszke2019}, a widely used deep learning framework that provides dynamic computation graphs and efficient GPU acceleration. Training was performed on a workstation equipped with an NVIDIA RTX2070 GPU to expedite the computational process. Further details, if required, are provided in Table \ref{tab:NN_Architecture_Summary} in \ref{app3}.

\paragraph{Computational Efficiency} 

Compared to the traditional A* algorithms, the neural network offers significant computational speed-ups due to its feedforward nature, eliminating the need for iterative graph traversal. Additionally, computation of the energy cost over the trajectory is not required - like in the A* planners - reducing the computational hardware requirement and energy consumption. This efficiency is particularly advantageous for real-time applications where computational resources are limited and thus why it has been selected for analysis in this work.

\subsection{Limitations and Considerations} 

While the neural network provides rapid trajectory predictions, it inherently relies on the quality and diversity of the training data. As the model is trained on trajectories generated by the A* planners, its performance is directly tied to the scenarios represented in the training set. Extrapolation to unseen flow conditions or starting positions not represented in the training data may result in decreased accuracy, and will be assessed within this analysis.

\section{Results} 
\label{Results} 

This section presents a comparative performance analysis of the four path planning approaches developed and investigated in this study. These are:

\begin{enumerate}
    \item Current-informed A* planner ($C.I._{A*}$)
    \item Wake-informed A* planner ($W.I._{A*}$)
    \item Current-informed neural network ($C.I._{NN}$)
    \item Wake-informed neural network ($W.I.{NN}$)
\end{enumerate}

The performance of each approach is evaluated based on several key metrics. These are the total energy expenditure of the generated paths ($E$) in Table \ref{tab:ResultsTable1}, the total path length ($L$) in meters in Table \ref{tab:ResultsTable2}, the number of high velocity "risky" cells that the paths intersect ($N_{\text{high-velocity}}(P)$) in Table \ref{tab:ResultsTable3}, the number of turbulent "risky" cells that the paths intersect ($N_{\text{turbulent}}(P)$) in Table \ref{tab:ResultsTable3} and the time taken to generate a given path ($t$) in Table \ref{tab:ResultsTable4}.

The aggregated results (mean and standard deviation across speed and angle combinations) are presented in Tables \ref{tab:ResultsTable1} - \ref{tab:ResultsTable4}, where the optimal values have been highlighted. Figures \ref{fig:std_energy}, \ref{fig:median_length}, and \ref{fig:median_highvelocity} provide further visual analysis of the results' distribution and trends against flow speed and angle.

\subsection{Energy Expenditure} 

Energy efficiency is critical in AUV operation due to their limited onboard energy resources. The wake-informed A* planner ($W.I._{A*}$) consistently demonstrates the lowest total energy expenditure, as determined via Eq. \ref{eqn:PathEnergy}, across all speed ranges, as shown in Table \ref{tab:ResultsTable1}. At the lowest speed range of 0.1--0.5m/s, $W.I._{A*}$ achieves an average energy expenditure of 51.49 joules with a standard deviation of 52.82 joules, outperforming the current-informed A* planner ($C.I._{A*}$) by approximately 11.3\%. This persists at higher speeds, with $W.I._{A*}$ maintaining lower energy consumption compared to the other models. At the maximum tested speed range of 4.0--5.0m/s, $W.I._{A*}$ records an average energy expenditure of 11,952.30 joules,  7.2\% less than $C.I._{A*}$.

\begin{figure*}[!ht]
    \centering
    \includegraphics[width=0.85\textwidth]{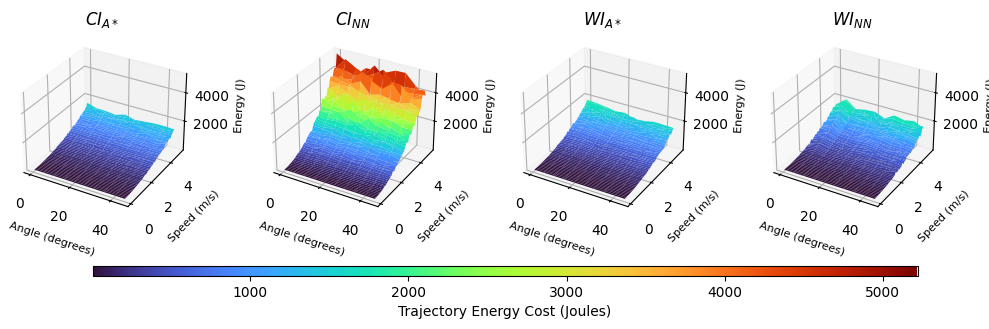}
    \caption{Standard deviation of the energy expenditure against speed and inflow angle for the current-informed A* path planner ($C.I._{A*}$), current-informed neural network ($C.I._{NN}$), wake-informed A* path planner ($W.I._{A*}$) and wake-informed neural network ($W.I._{NN}$) }
    \label{fig:std_energy}
\end{figure*}

The neural network models, $C.I._{NN}$ and $W.I._{NN}$, based on their counterpart A* implementations, exhibit higher energy expenditures compared to their calculated A* trajectories. $W.I._{NN}$ performs better than $C.I.{NN}$, but does not match the energy efficiency of the wake-informed A* planner. The increased energy expenditure (ranging between 10.03--13.12\%) in the neural network counterparts may be attributed to approximation errors inherent in learning-based approaches, where it remains challenging to capture the optimality of the A* algorithm in path planning, especially in the presence of complex wake structures. Provided in Fig. \ref{fig:std_energy} is a depiction of the variation in standard deviation against speed and angle for both planners and networks. It can be seen that the current-informed NN has significant variation in path energy, indicating it may struggle to understand how best to traverse the domain, particularly at high speeds.

\subsection{Path Length} 

\begin{figure*}[!ht]
    \centering
    \includegraphics[width=0.85\textwidth]{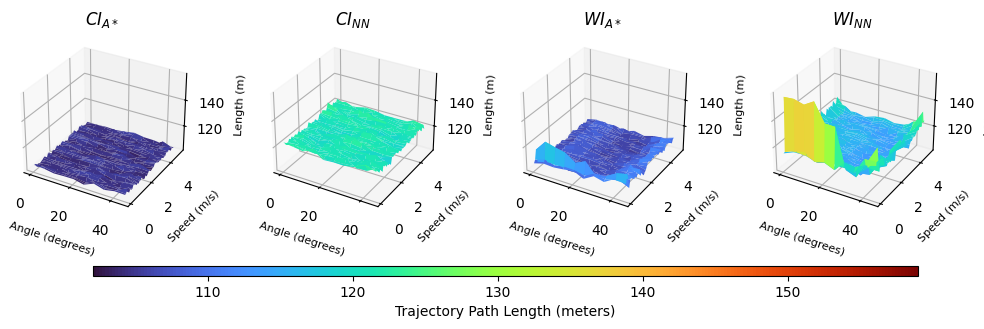}
    \caption{Median trajectory length against speed and inflow angle for the current-informed A* path planner ($C.I._{A*}$), current-informed neural network ($C.I._{NN}$), wake-informed A* path planner ($W.I._{A*}$) and wake-informed neural network ($W.I._{NN}$) }
    \label{fig:median_length}
\end{figure*}

The path length metric ($L$), as given in Table \ref{tab:ResultsTable2}, and determined via Eq. (\ref{eqn:PathLength}), reflects the total distance traveled by the AUV along the planned trajectory. Shorter paths generally result in quicker LAR missions, for a given speed, but may traverse more hazardous areas. The current-informed A* planner ($C.I._{A*}$) achieves the shortest average path lengths across all speed ranges, with consistent trajectory lengths of  104.22--104.31m and a standard deviation of approximately 9.5m. This suggests that $C.I._{A*}$ is able to generate the most direct paths, but without considering wake structures and associated hazards. In contrast, the wake-informed A* planner ($W.I._{A*}$) shows slightly longer path lengths, averaging around 107 m, indicating that it deliberately avoids certain areas to minimize energy expenditure and exposure to high-velocity regions. The neural network models, particularly $C.I._{NN}$, exhibit significantly longer path lengths with higher variability. For example, $C.I._{NN}$ records an average path length of 133.27m in the 4.0--5.0m/s speed range, approximately 27\% longer than $C.I._{A*}$. This suggests that the neural networks may not approximate the optimal path lengths as effectively, potentially leading to suboptimal routing. Fig. \ref{fig:median_length} provides an assessment of the median path length of the models against speed and angle. It can be seen that the A* current-informed method is the most consistent, with the A* wake-informed method also exhibiting similar performance. The neural networks result in a higher median path length, with suboptimal path generation at low speed and angles for the wake-informed NN.

\subsection{High-Velocity Cells Encountered} 

\begin{figure*}[!ht]
    \centering
    \includegraphics[width=0.85\textwidth]{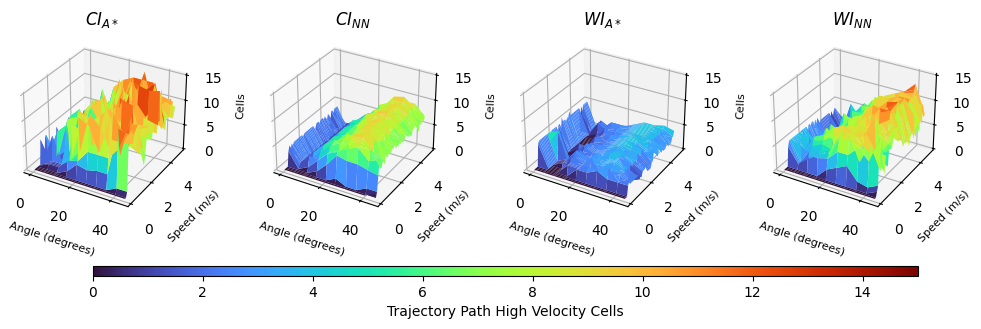}
    \caption{Median count of the trajectories' encountered high velocity cells ($N_{\text{high-velocity}}(P)$) against speed and inflow angle for the current-informed A* path planner ($C.I._{A*}$), current-informed neural network ($C.I._{NN}$), wake-informed A* path planner ($W.I._{A*}$) and wake-informed neural network ($W.I._{NN}$) }
    \label{fig:median_highvelocity}
\end{figure*}

Navigating through high-velocity cells poses increased risk and potential for energy expenditure, due to increased drag and control challenges associated with traversing the larger vehicle's wake structure. This data is provided in Table \ref{tab:ResultsTable3}. The wake-informed A* planner ($W.I._{A*}$) exhibits the lowest average number of high-velocity cells encountered ($N_{\text{high-velocity}}(P)$) across all speed ranges, as it has knowledge of the structures. At speeds between 0.5--1.0~m/s, $W.I._{A*}$ encounters an average of 2.73 high-velocity cells, significantly lower than $C.I._{A*}$ (7.32) on average. This indicates that $W.I._{A*}$ sufficiently avoids the high-velocity regions it is informed of, enhancing operational safety. Both neural network models exhibit less consistent performance in this metric. $C.I._{NN}$ is observed to reduce high-velocity cell encounters compared to $C.I._{A*}$, whilst $W.I._{NN}$ generally encounters more high-velocity cells than $W.I._{A*}$. This suggests that the neural networks may not fully capture the nuances of avoiding hazardous regions, possibly due to limitations in the training data or model capacity. Fig. \ref{fig:median_highvelocity} details the median number of high velocity cells encountered by each method. Similar performance is observed for the current-informed A* and NN, and the wake-informed NN. The wake-informed A* planner exhibits beneficial performance in comparison to the other models, with typically 0 - 5 cells encountered irrespective of speed and angle. 

\subsection{Turbulent Cell Count} 

The turbulent cell count ($N_{\text{turbulent}}(P)$), as given in Table \ref{tab:ResultsTable3}, reflects the number of cells along the trajectory where any velocity fluctuation occurs, which would pose an additional control challenge for the AUV. It should be noted that no thresholding is applied to this, so small and large fluctuations are treated equally. The current-informed A* planner ($C.I._{A*}$) records the lowest turbulent cell counts across all speed ranges, with an average of 31.29 cells in the 0.5--1.0~m/s range. The wake-informed models, both $W.I._{A*}$ and $W.I._{NN}$, exhibit higher turbulent cell counts. This suggests that while the wake-informed planners effectively avoid high-velocity cells, they may traverse areas with higher turbulence. One possible explanation is that in avoiding high-velocity wake regions, the wake-informed planners route the AUV through areas with greater velocity disparities. The neural network models again display higher turbulent cell counts compared to their A* counterparts, indicating a potential area for improvement in modeling turbulence avoidance.

\subsection{Computational Time} 

Computational efficiency is critical for real-time path planning on resource-constrained AUVs. As shown in Table \ref{tab:ResultsTable4}, the neural network models offer a significant advantage in computational time. The neural networks compute trajectories in approximately 0.30 milliseconds, whereas the A* planners require 37 to 43 seconds on average, depending on the method. This is a compute time reduction of approximately 5 orders of magnitude - a significant decrease. The wake-informed NN, $W.I._{NN}$, achieves the shortest computation time across all speed ranges, with an average of 0.29 milliseconds. The substantial reduction in computation time makes the neural network models highly optimal for real-time applications. This speed, however, comes at the cost of reduced path optimality compared to direct computation via A*, as evidenced by the higher energy expenditures and longer path lengths compared to both A* planners.

\begin{table*}[!ht]
\centering
\begin{tabular}{ l l l l l l } \hline 
  Metric [$ \downarrow $] & Range [m/s] & $C.I._{A*}$ & $C.I._{NN}$ & $W.I._{A*}$ & $W.I._{NN}$  \\ \hline 
  $E$ $[J]$     & 0.1 - 0.5 & 58.04 $\pm$ 56.76      & 64.17 $\pm$ 67.25      & \textbf{51.49 $\pm$ 52.82}      & 58.72 $\pm$ 58.19\\ 
                & 0.5 - 1.0 & 406.17 $\pm$ 146.48    & 455.76 $\pm$ 202.19    & \textbf{375.68 $\pm$ 138.67}    & 414.85 $\pm$ 153.06\\
                & 1.0 - 2.0 & 1537.47 $\pm$ 573.90   & 1728.46 $\pm$ 784.70   & \textbf{1428.01 $\pm$ 543.16}   & 1582.76 $\pm$ 601.81 \\
                & 2.0 - 3.0 & 4074.23 $\pm$ 984.63   & 4581.10 $\pm$ 1588.40  & \textbf{3776.55 $\pm$ 950.04}   & 4233.10 $\pm$ 1047.98\\
                & 3.0 - 4.0 & 7861.94 $\pm$ 1458.17  & 8863.53 $\pm$ 2731.18  & \textbf{7304.46 $\pm$ 1436.68}  & 8115.69 $\pm$ 1556.69\\
                & 4.0 - 5.0 & 12873.20 $\pm$ 2021.45 & 14578.18 $\pm$ 4253.80 & \textbf{11952.30 $\pm$ 2046.57} & 13467.80 $\pm$ 2358.98\\ 
  \hline
\end{tabular}
\caption{Energy expenditure (Eq. (\ref{eqn:PathEnergy})) denoted by $E$ for the A* planners and neural networks. Mean and standard deviation provided for the current-informed A* path planner ($C.I._{A*}$), current-informed neural network ($C.I._{NN}$), wake-informed A* path planner ($W.I._{A*}$) and wake-informed neural network ($W.I._{NN}$).}
\label{tab:ResultsTable1}
\end{table*}

\begin{table*}[!h]
\centering
\begin{tabular}{ l l  l l l l } \hline 
  Metric [$ \downarrow $] & Range [m/s] & $C.I._{A*}$ & $C.I._{NN}$ & $W.I._{A*}$ & $W.I._{NN}$  \\ \hline 
  $L$ $[m]$     & 0.1 - 0.5 & \textbf{104.29 $\pm$ 9.67}      & 133.18 $\pm$ 30.30     & 111.51 $\pm$ 12.22     & 125.81 $\pm$ 23.26\\ 
                & 0.5 - 1.0 & \textbf{104.25 $\pm$ 9.55}      & 132.51 $\pm$ 30.58     & 107.72 $\pm$ 9.46      & 116.77 $\pm$ 11.14\\
                & 1.0 - 2.0 & \textbf{104.26 $\pm$ 9.47}      & 132.44 $\pm$ 30.41     & 107.27 $\pm$ 9.33      & 116.34 $\pm$ 10.38\\
                & 2.0 - 3.0 & \textbf{104.22 $\pm$ 9.49}      & 132.45 $\pm$ 30.42     & 107.32 $\pm$ 9.50      & 117.60 $\pm$ 10.85\\
                & 3.0 - 4.0 & \textbf{104.31 $\pm$ 9.55}      & 132.78 $\pm$ 30.54     & 107.17 $\pm$ 9.42      & 116.91 $\pm$ 10.12\\
                & 4.0 - 5.0 & \textbf{104.30 $\pm$ 9.57}      & 133.27 $\pm$ 30.65     & 107.22 $\pm$ 9.50      & 118.28 $\pm$ 10.88\\ 
  \hline
\end{tabular}
\caption{Path length (Eq. \ref{eqn:PathLength}) in meters denoted by $L$ for the A* planners and neural networks. Mean and standard deviation provided for the current-informed A* path planner ($C.I._{A*}$), current-informed neural network ($C.I._{NN}$), wake-informed A* path planner ($W.I._{A*}$) and wake-informed neural network ($W.I._{NN}$).}
\label{tab:ResultsTable2}
\end{table*}

\begin{table*}[!ht]
\centering
\begin{tabular}{ l l  l l l l } \hline 
  Metric [$ \downarrow $] & Range [m/s] & $C.I._{A*}$ & $C.I._{NN}$ & $W.I._{A*}$ & $W.I._{NN}$  \\ \hline 
                
  $N_{\text{high-velocity}}(P)$   & 0.1 - 0.5 & 2.27 $\pm$ 4.30         & 2.28 $\pm$ 3.45         & \textbf{0.83 $\pm$ 1.41}         & 2.18 $\pm$ 3.71\\
               & 0.5 - 1.0 & 7.32 $\pm$ 5.96         & 6.00 $\pm$ 4.80         & \textbf{2.73 $\pm$ 1.58}         & 6.97 $\pm$ 5.21\\
                & 1.0 - 2.0 & 7.31 $\pm$ 6.03         & 6.29 $\pm$ 4.85         & \textbf{2.91 $\pm$ 1.71}         & 7.31 $\pm$ 5.00\\
                & 2.0 - 3.0 & 7.32 $\pm$ 6.02         & 6.22 $\pm$ 4.97         & \textbf{2.81 $\pm$ 1.64}         & 7.43 $\pm$ 5.26\\
                & 3.0 - 4.0 & 7.49 $\pm$ 6.01         & 6.24 $\pm$ 5.06         & \textbf{3.01 $\pm$ 1.74}         & 7.24 $\pm$ 5.49\\
                & 4.0 - 5.0 & 7.38 $\pm$ 6.09         & 6.22 $\pm$ 5.13         & \textbf{3.06 $\pm$ 1.75}         & 7.25 $\pm$ 5.32\\ \hline
                
  $N_{\text{turbulent}}(P)$   & 0.1 - 0.5 & \textbf{54.19 $\pm$ 19.16}       & 60.49 $\pm$ 17.00       & 66.14 $\pm$ 19.90         & 65.02 $\pm$ 19.36\\
                 & 0.5 - 1.0 & \textbf{31.29 $\pm$ 12.60}       & 41.47 $\pm$ 12.70       & 47.65 $\pm$ 16.11         & 45.71 $\pm$ 14.23\\
                & 1.0 - 2.0 & \textbf{29.69 $\pm$ 12.65}       & 39.84 $\pm$ 11.94       & 44.90 $\pm$ 14.51         & 44.10 $\pm$ 12.96\\
                & 2.0 - 3.0 & \textbf{33.12 $\pm$ 13.92}       & 42.29 $\pm$ 12.12       & 46.82 $\pm$ 14.62         & 46.56 $\pm$ 13.19\\
                & 3.0 - 4.0 & \textbf{35.14 $\pm$ 13.84}       & 44.29 $\pm$ 12.20       & 47.79 $\pm$ 14.16         & 47.63 $\pm$ 12.93\\
                & 4.0 - 5.0 & \textbf{37.04 $\pm$ 14.54}       & 46.26 $\pm$ 12.54       & 49.47 $\pm$ 14.27         & 49.02 $\pm$ 13.11\\ 
  \hline
\end{tabular}
\caption{Number of high velocity cells $N_{\text{high-velocity}}(P)$ and the turbulent cell count $N_{\text{turbulent}}(P)$ (Eq. \ref{eqn:PathHighVelocity} \& \ref{eqn:PathTurbulence}), encountered on a given trajectory $P$ by the A* planners and neural networks. Mean and standard deviation provided for the current-informed A* path planner ($C.I._{A*}$), current-informed neural network ($C.I._{NN}$), wake-informed A* path planner ($W.I._{A*}$) and wake-informed neural network ($W.I._{NN}$).}
\label{tab:ResultsTable3}
\end{table*}

\begin{table*}[!ht]
\centering
\begin{tabular}{ l l l l l l } \hline 
  Metric [$ \downarrow $] & Range [m/s] & $C.I._{A*}$ & $C.I._{NN}$ & $W.I._{A*}$ & $W.I._{NN}$  \\ \hline 
  $t$ $[sec]$   & 0.1 - 0.5 & 42.81 $\pm$ 10.41 & 0.00034 $\pm$ 0.05308 & 26.46 $\pm$ 222.35 & \textbf{0.00029 $\pm$ 0.00016} \\
                & 0.5 - 1.0 & 42.81 $\pm$ 10.41 & 0.00033 $\pm$ 0.00012 & 39.81 $\pm$ 15.08 & \textbf{0.00028 $\pm$ 0.00014} \\
                & 1.0 - 2.0 & 42.81 $\pm$ 10.40 & 0.00035 $\pm$ 0.00020 & 39.66 $\pm$ 13.93 & \textbf{0.00029 $\pm$ 0.00018} \\
                & 2.0 - 3.0 & 42.81 $\pm$ 10.40 & 0.00038 $\pm$ 0.00045 & 37.79 $\pm$ 13.05 & \textbf{0.00031 $\pm$ 0.00049} \\
                & 3.0 - 4.0 & 42.81 $\pm$ 10.40 & 0.00034 $\pm$ 0.00049 & 38.89 $\pm$ 13.05 & \textbf{0.00029 $\pm$ 0.00025} \\
                & 4.0 - 5.0 & 42.81 $\pm$ 10.40 & 0.00035 $\pm$ 0.00026 & 37.17 $\pm$ 12.82 & \textbf{0.00031 $\pm$ 0.00023} \\ 
  \hline
\end{tabular}
\caption{Computational time for a given path by the A* planners and neural networks. Mean and standard deviation provided for the current-informed A* path planner ($C.I._{A*}$), current-informed neural network ($C.I._{NN}$), wake-informed A* path planner ($W.I._{A*}$) and wake-informed neural network ($W.I._{NN}$). }
\label{tab:ResultsTable4}
\end{table*}

\section{Discussion} 
\label{Discussion}



The results demonstrate a trade-off between computational efficiency and path optimality. Knowledge of the wake, via the wake-informed A* planner ($W.I._{A*}$), provides the most energy-efficient and safe paths but requires significant computational resources and time. Conversely, the neural network models offer near-instantaneous computation times, critical for real-time operations, however, the resulting paths are less optimal in terms of energy consumption and safety metrics, which may hinder their suitability. These findings highlight the importance of incorporating detailed environmental information, such as wake structures, into path planning algorithms. Although neural networks capture some aspects of the optimal paths, there is room for improvement in their ability to approximate the solutions provided by the A* planners.

Previous studies have demonstrated the effectiveness of neural networks in path planning tasks for autonomous systems \citep{Tai2017, Pfeiffer2017}. However, most of these approaches focus on simplified environments or two-dimensional spaces. Our results extend this understanding by evaluating and demonstrating the potential of using neural network-based planners in complex three-dimensional underwater environments with dynamic wake effects.  

The observed reduction in path optimality (higher energy cost, longer paths) when using the NN approximations carries significant implications. For missions operating under tight energy budgets or requiring precise maneuvering near hazards, the 4.51--19.79\% increase in energy expenditure or deviations from the truly optimal path might be unacceptable. This highlights a critical challenge: while NNs offer the computational speed needed for real-time reactivity, their current level of precision might limit their applicability in safety- or efficiency-critical scenarios without further safeguards or improvements. The approximation errors likely arise from the inherent difficulty for a feedforward network to perfectly replicate the complex, iterative decision-making process of A* within a high-dimensional, dynamic 3D flow field, especially given the potentially non-linear relationships between inputs and optimal outputs.

While the dramatic reduction in computation time is the most prominent advantage of the NN approach demonstrated here, other potential benefits, do exist. NNs offer a fixed, predictable inference time once trained, unlike A* whose runtime can vary considerably depending on the specific start/goal configuration and environment complexity. If trained on a sufficiently diverse dataset encompassing various environmental conditions and vehicle states, NNs might exhibit smoother interpolation and potentially better generalization to unforeseen (but similar) scenarios compared to A*'s discrete grid search. It should be noted that this adaption would requires rigorous validation against out-of-distribution data, potentially from a much larger set of paths $P$. Furthermore, the NN framework is inherently adaptable; it could potentially incorporate diverse real-time sensor inputs (e.g., sonar, Doppler Velocity Log) beyond the pre-defined parameters used here, allowing for more reactive planning based on immediate perception -- and potentially tailored to specific vehicle configurations. Although not explicitly optimized for here, certain NN architectures could potentially learn to generate inherently smoother paths, which might be more dynamically feasible for an AUV to follow, reducing control effort compared to potentially jagged A* paths on a grid. However, based on the results here, comparing the optimality metrics, it is the computational speed, compute hardware and energy requirement reduction that remains the primary realized advantages of the NN approach in this work.

Deploying either the A* or NN path planners in real-world AUV operations involves significant practical complications beyond the idealized simulation environment. For A*, the primary hurdles are its computational demand which challenges real-time re-planning on typical AUV onboard processors, especially in dynamic environments, and its reliance on a pre-defined, accurate environmental map. Real-world sensor data (e.g., sonar for obstacle detection, ADCP for currents) is often noisy, sparse, and provides incomplete coverage, making the creation of a reliable, up-to-date grid map difficult. Furthermore, the discretization of the continuous world into a grid can itself introduce sub-optimality or miss feasible paths. 

For NNs, despite their online speed, the main concerns are reliability and safety. Their performance is fundamentally tied to the quality and representativeness of the training data. Encountering real-world conditions significantly different from the training scenarios (out-of-distribution problem) can lead to unpredictable and potentially unsafe behavior. Verification and validation of NN controllers for safety-critical applications like AUV navigation remain open research challenges. Both methods, as implemented here, also simplify AUV dynamics, neglecting constraints like minimum turning radius, acceleration/deceleration limits, and control response times, which are crucial for path feasibility. Additionally, handling unforeseen dynamic obstacles (e.g., marine life, other vessels) requires integration with reactive collision avoidance systems, which is not covered in the current planning framework. Bridging the gap between these simulation-based planners and robust real-world deployment requires addressing sensor integration, uncertainty management, dynamic replanning, vehicle constraints, and rigorous safety validation.

Several avenues exist for potentially improving the precision and reliability of NN-based planners in this context. Firstly, exploring more sophisticated network architectures tailored to sequential decision-making, such as Recurrent Neural Networks (RNNs, e.g., LSTMs) or Transformers \citep{Vaswani2017}, could better capture the temporal dependencies along the path and potentially model the environmental interactions more effectively. Secondly, integrating physics knowledge directly into the network, perhaps through Physics-Informed Neural Networks (PINNs) \citep{Raissi2019} or by designing loss functions that explicitly penalize physically implausible or high-energy path segments, might constrain the network to produce more realistic and efficient trajectories. Thirdly, enhancing the training strategy, for instance by using Generative Adversarial Networks (GANs) to generate paths or employing reinforcement learning (RL) where the agent learns a policy directly by interacting with the environment or a simulator, could lead to better performance, although RL often requires substantial training time and careful reward shaping. Finally, hybrid approaches, where an NN provides a fast initial path suggestion that is then locally refined using a quicker optimization or search algorithm (like a limited-depth A* or potential field method), could offer a pragmatic balance between speed and near-optimality.

The significant computational speed-ups offered by the neural networks, are well understood. This work shows that the extension to 3-dimensional planning aligns with expected findings in the current state of the literature~\citep{LeCun2015}. The observed reductions in path optimality, however, also suggest that further advancements in neural network architectures or training methodologies could be beneficial to close the performance gap with traditional A* algorithms in complex, highly dynamic environments.

\section{Conclusions} 
\label{Conclusions}

This study demonstrated the substantial benefits of incorporating detailed 3D wake structures into AUV path planning. The wake-informed A* algorithm ($W.I._{A*}$) consistently yielded the most energy-efficient paths, reducing energy consumption by up to 11.3\% compared to its current-informed counterpart ($C.I._{A*}$) while also minimizing encounters with high-velocity wake regions. Neural network approximations ($C.I._{NN}$ and $W.I._{NN}$) achieved dramatic computational speed-ups of approximately six orders of magnitude over A*, offering a viable path towards real-time implementation. However, this speed came at the cost of reduced path optimality, with NNs exhibiting 4.5--19.8\% higher energy expenditures and generating 9.8--24.4\% longer paths. These core findings highlight the critical trade-off between path optimality achievable with detailed environmental information and the computational feasibility required for onboard, real-time AUV operations.

Future research should prioritize bridging the performance gap between the rapid NN approximations and the optimal A* solutions, while also addressing real-world applicability. Key directions include: exploring advanced neural network architectures (e.g., RNNs, Transformers, physics-informed models) better suited for sequential and physics-based tasks; developing hybrid planning systems that leverage NN speed for initial path generation followed by algorithmic refinement; explicitly incorporating AUV dynamic constraints and control limitations into the planning process; and establishing robust methods for uncertainty quantification and safety validation of learning-based planners in complex, dynamic underwater environments. Progress in these areas is essential for enabling safe, efficient, and autonomous AUV operations in close-proximity scenarios characterized by significant hydrodynamic interactions.

\appendix 

\section*{Nomenclature}
\label{app0}

\begin{table}[!ht]
    \small
    \centering
    \begin{tabular}{c l}
    \hline
    $G$                 &  3D search grid \\
    $n$                 &  Given node in trajectory \\
    $n'$                &  Neighboring node in trajectory \\
    $N$                 &  Number of nodes\\
    $d$                 &  Euclidean distance \\
    $F_{D}$             &  Force of drag\\
    $\rho$              & Density \\
    $v$                 & Velocity \\
    $C_{D}$             & Coefficient of drag  \\
    $A$                 & Surface area  \\
    $E_{D}$             & Energy cost over distance \\
    $\omega$            & Weighting factor \\
    $c(n)$              & Cost \\
    $g(n)$              & g-score \\
    $P$                 &  Path of $n$ nodes\\
    $\mathcal{P}$       &  Set of possible paths \\
    $h(n)$              & Search heuristic \\
    $P^*_{C}$           & Current-informed planner \\
    $P^*_{W}$           & Wake-informed planner \\
    $E$                 & Energy (joules)\\
    $L$                 & Length (meters)\\
    $N_{\text{high-velocity}}(P)$ & High velocity cells \\
    $N_{\text{turbulent}}(P)$ & Turbulent cells\\
    $\sigma_v$          & Velocity standard deviation in $G$\\
    $\overline{v}$      & Median velocity in $G$\\
    $\delta_{v}$        & Indicator function (velocity)\\
    $\delta_{T}$        & Indicator function (turbulence)\\
    $\mathcal{L}$       & Training loss \\
    $y_i$, $\hat{y}_i$  & True, predicted path coordinates\\
    $m_i$               & Coordinate path mask \\
    $C.I._{A*}$         &  Current-informed A* planner\\ 
    $C.I._{NN}$         &  Current-informed neural network\\ 
    $W.I._{A*}$         &  Wake-informed A* planner\\ 
    $W.I._{NN}$         &  Wake-informed neural network\\
    \hline
    \end{tabular}
\end{table}

\section{Dataset details}
\label{app1}

The wake flow field data, as discussed in \ref{Path Generation}, was generated from computational fluid dynamics (CFD) calculations undertaken using the commercial software package ANSYS Fluent 2021R1. The flow fields were generated using  Reynolds Averaged Navier-Stokes (RANS) simulations with the $k-\epsilon$ turbulence model. An XLUUV geometry based on \citep{IEEE} was located in the center of a domain, sized to be 5L (with a characteristic length $L$ of 22 meters) in each axial direction, to ensure boundary effects were mitigated. The $y^{+}$ of the simulation was ensured to be in the range $30 < y^{+} < 150$, to ensure applicability of the $k-\epsilon$ turbulence model. An INSEAN E1619 propeller model was used, based on existing literature \citep{SUBSTEC, IEEE}. The domain was discretized into 15 million cells, with the majority of these cells clustered in the wake region of the XLUUV using the local refinement method of bodies of influence from \citep{IEEE}. The freestream velocity was varied from 0.10 m/s to 5.00 m/s in increments of 0.10 m/s and angles of incidence from 0° to 60° degrees of separation in increments of 5°. For each speed and angle pair, a RANS simulation was computed until residual convergence of 1E-5 was achieved. The data from these CFD meshes was then interpolated onto the grid ($G$), using nearest neighbor interpolation from the mesh cells to the cells of the $128^3$ grid. The final dataset of distinct grids comprised 500 speed and angle CFD fields. This dataset can be made available upon request.

\section{Vehicle details}
\label{app2}

Table \ref{tab:AUV_Particulars} details the AUV model used for the development of the energy cost (Eq. \ref{eqn:DragEnergy}). Provided are the physical design it was modeled on, the length, diameter, cross sectional area, drag coefficient and the water density.

\begin{table*}[!ht]
    \centering
    \begin{tabular}{lll}
    \hline 
     Parameter & Value & Source \\
    \hline 
    AUV Model            & GRAALTech X300 & \citep{SUBSTEC, AUV_REF} \\
    AUV Length           & 3.060 $m$ & \citep{AUV_REF} \\ 
    AUV Diameter         & 0.254 $m$ & \citep{AUV_REF} \\ 
    AUV Cross sectional area (A) & 0.051 $m^2$ & \citep{AUV_REF} \\ 
    AUV Drag Coefficient & 0.15 &  \citep{AUV_REF} \\ 
    Water density & 1025.1627 $kg/m^{3}$  &  \citep{IEEE} \\ \hline 
    \end{tabular}
    \caption{Physical particulars of the AUV platform used to guide the path planner A* methods and inform the energy compute as listed in Section \ref{Path Planner Operation}}
    \label{tab:AUV_Particulars}
\end{table*}

\section{NN Implementation}
\label{app3}

Table \ref{tab:NN_Architecture_Summary} provides the general structure, input, layer, output, training design and build of the two NN variants used to replicate the trajectories of the current-informed A* and wake-informed A* planners. These specifications are provided to ensure the work can be replicated. Access to our code repository can be made available upon request.
 
\begin{table*}[!ht]
    \centering
    \caption{Summary of Neural Network Architecture and Training Configuration.}
    \label{tab:NN_Architecture_Summary}
    \begin{tabular}{L{1.5cm} L{2.75cm} L{11.75cm}}
        \hline
        Category          & Item                     & Specification \\
        \hline
        General           & Network Type             & Fully Connected Feedforward Neural Network \\
                          & Purpose                  & Regress sequence of 3D waypoints from input parameters \\
                          & Variants                 & Two identical networks trained independently ($C.I._{NN}$, $W.I._{NN}$) \vspace{0.3cm} \\
        Input             & Dimension                & 8 \\
                          & Features                 & Start Coords ($x_s, y_s, z_s$), Goal Coords ($x_g, y_g, z_g$), Flow Speed ($v$), Flow Angle ($\theta$) \\
                          & Preprocessing            & Normalization (Z-score based on training set mean/std, Eq. \ref{eqn:StandardScore}) \vspace{0.3cm} \\
        Layers            & Layer 1 (Input)          & Linear (Dense), Input: 8, Output: 128 \\
                          & Activation 1             & ReLU ($\max(0, x)$) \\
                          & Layer 2 (Hidden)         & Linear (Dense), Input: 128, Output: 256 \\
                          & Activation 2             & ReLU ($\max(0, x)$) \\
                          & Layer 3 (Output)         & Linear (Dense), Input: 256, Output: 390 \vspace{0.3cm} \\
        Output            & Dimension                & 390 \\
                          & Structure                & Flattened sequence of 130 waypoints: $(x_1, y_1, z_1, \dots, x_{130}, y_{130}, z_{130})$ \\
                          & Postprocessing           & Denormalization (using training set mean/std) \\
                          & Length                   & Output padded to 130 waypoints; Mask vector used in loss \vspace{0.3cm} \\
        Training          & Training Data            & 14,400 trajectories (80\% of A* generated paths) \\
                          & Validation Data          & 3,600 trajectories (20\% of A* generated paths) \\
                          & Loss Function            & Masked Mean Squared Error (MSE, Eq. \ref{eqn:MaskedMSE}) \\
                          & Optimizer                & Adam \citep{Kingma2015} \\
                          & Learning Rate            & $1 \times 10^{-4}$ \\
                          & Batch Size               & 64 \\
                          & Max Epochs               & 50 \\
                          & Initialization           & Xavier / Glorot \citep{Glorot2010} \\
                          & Regularization           & Early Stopping (Patience: 10 epochs on validation loss) \vspace{0.3cm} \\
        Build & Framework         & PyTorch \citep{Paszke2019} \\
        \hline 
    \end{tabular}
\end{table*}




\bibliographystyle{elsarticle-harv} 
\bibliography{bibliography}






\end{document}